\documentclass[10pt,twocolumn,letterpaper]{article}

\usepackage{cvm}
\usepackage{times}
\usepackage{epsfig}
\usepackage{graphicx}
\usepackage{amsmath}
\usepackage{amssymb}
\usepackage{pifont}  
\usepackage{multirow}
\usepackage{xcolor}
\newcommand{\equalcontribution}{\textsuperscript{*}}

\usepackage[pagebackref=true,breaklinks=true,colorlinks,bookmarks=false]{hyperref}
\newcommand{\keywords}[1]{{\bf \emph{Keywords: #1}}}

\cvmfinalcopy

\def\cvmPaperID{****} 

\ifcvmfinal\fi
\begin{document}

\title{Takin-ADA: Emotion Controllable Audio-Driven Animation with Canonical and Landmark Loss Optimization}
\author{Bin Lin\equalcontribution\\ 
Ximalaya Inc., ShangHai, China\\
{\tt\small bin.lin@ximalaya.com}
\and
Yanzhen Yu\equalcontribution\\
Ximalaya Inc., ShangHai, China\\
{\tt\small yanzhen.yu@ximalaya.com}
\and
Jianhao Ye$^\dagger$\\
Ximalaya Inc., ShangHai, China\\
{\tt\small jianhao.ye@ximalaya.com}
\and
Ruitao Lv\\
Ximalaya Inc., ShangHai, China\\
{\tt\small ruitao.lv@ximalaya.com}
\and
Yuguang Yang\\
Ximalaya Inc., ShangHai, China\\
{\tt\small yuguang.yang@ximalaya.com}
\and
Ruoye Xie\\
Ximalaya Inc., ShangHai, China\\
{\tt\small ruoye.xie@ximalaya.com}
\and
Pan Yu\\
Ximalaya Inc., ShangHai, China\\
{\tt\small yu2.pan@ximalaya.com}
\and
Hongbin Zhou\\
Ximalaya Inc., ShangHai, China\\
{\tt\small hongbin.zhou@ximalaya.com}
}
\maketitle
\begin{abstract}
   
    \vspace{-3mm}
    Existing audio-driven facial animation methods face critical challenges, including expression leakage, ineffective subtle expression transfer, and imprecise audio-driven synchronization. We discovered that these issues stem from limitations in motion representation and the lack of fine-grained control over facial expressions. To address these problems, we present \textit{Takin-ADA}, a novel two-stage approach for real-time audio-driven portrait animation. In the first stage, we introduce a specialized loss function that enhances subtle expression transfer while reducing unwanted expression leakage. The second stage utilizes an advanced audio processing technique to improve lip-sync accuracy. Our method not only generates precise lip movements but also allows flexible control over facial expressions and head motions. Takin-ADA achieves high-resolution (512x512) facial animations at up to 42 FPS on an RTX 4090 GPU, outperforming existing commercial solutions. Extensive experiments demonstrate that our model significantly surpasses previous methods in video quality, facial dynamics realism, and natural head movements, setting a new benchmark in the field of audio-driven facial animation.
    Examples are displayed on the demo page$^\ddagger$.
    \footnotetext[1]{These authors contributed equally to this work.}
    \footnotetext[2]{Corresponding author.}
    \footnotetext[3]{\url{https://everest-ai.github.io/takinada/}}
    
\end{abstract}


\begin{figure*}[ht]
    \centering
    \includegraphics[width=1\textwidth]{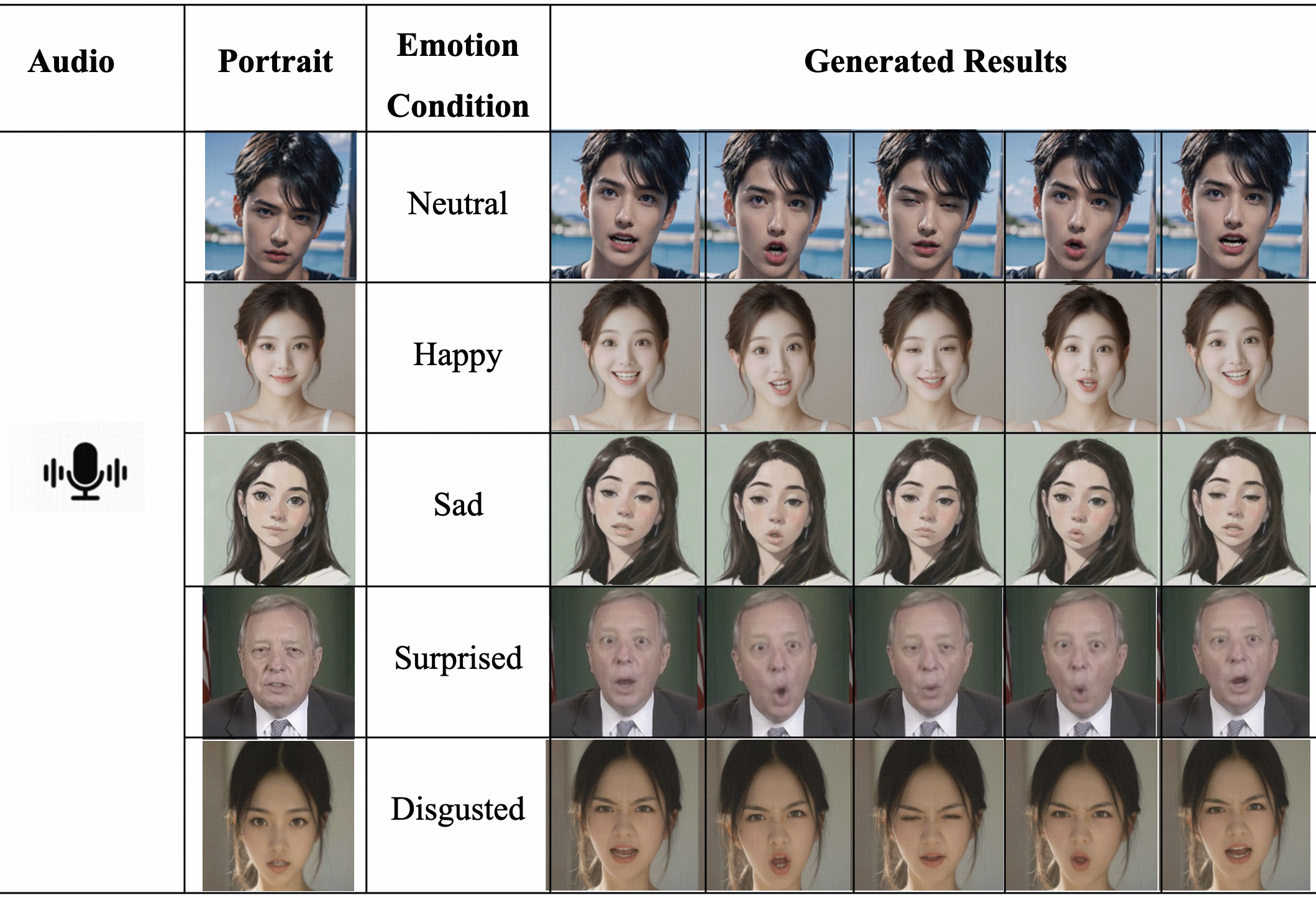}
    \caption{We introduce Takin-ADA, a framework that transforms input audio and a single static portrait into animated talking videos with naturally flowing movements. Each column of generated results utilizes identical control signals with different and expressions but incorporates some random variations, demonstrating the diversity of our generated outcomes.}
\end{figure*}

\section{Introduction}
In recent years, portrait animation has emerged as a pivotal area of research in computer vision, driven by its wide-ranging applications in digital human animation, film dubbing, and interactive media\cite{10.1145/3394171.3413532, hong2022depthawaregenerativeadversarialnetwork, zhong2023identitypreservingtalkingfacegeneration}. The ability to generate realistic, expressive, and controllable facial animations from a single image has become increasingly important in creating lifelike digital avatars for various applications, including virtual hosts, online education, and digital human interactions\cite{liu2024anitalkeranimatevividdiverse, xu2024vasa1lifelikeaudiodriventalking, 9577690}.

Existing approaches to portrait animation can be broadly categorized into two paradigms: audio-driven\cite{tan2024edtalkefficientdisentanglementemotional, 10.1145/3394171.3413532, zhong2023identitypreservingtalkingfacegeneration, zhang2021flow, zhou2021pose, zhou2020makelttalk} and video-driven animation\cite{wang2022latent, 9578110, guo2024liveportraitefficientportraitanimation}. While these methods have shown promise, they face significant challenges in achieving precise control over facial expressions, maintaining identity consistency, and generating natural head movements. Audio-driven methods often struggle to capture the full spectrum of non-verbal cues, resulting in animations that lack expressiveness\cite{Zhou_2020, 2021Audio2Head,yu2022talkingheadgenerationprobabilistic}. Video-driven techniques, while potentially capturing a wider range of facial dynamics, often suffer from expression leakage, where the source video's expressions unduly influence the animated output\cite{wang2022latent, tan2024edtalkefficientdisentanglementemotional}.

The primary challenge in this field lies in developing a unified framework that can simultaneously achieve individual facial control, handle both audio-driven and video-driven talking face generation efficiently, and operate in real-time. Existing models often rely on explicit structural representations such as blendshapes\cite{10094777, Fan_2022_CVPR, peng2023emotalk} or 3D Morphable Models (3DMM)\cite{danecek2022emocaemotiondrivenmonocular, 10.1145/3450626.3459936, ma2024dreamtalkemotionaltalkinghead}, which offer constrained approximations of facial dynamics and fail to capture the full breadth of human expressiveness.

To address these limitations, we present Takin-ADA (Audio-Driven Animation), an innovative two-stage framework for real-time audio-driven animation of single-image portraits with controllable expressions using 3D implicit keypoints\cite{9578110}. Our approach tackles the critical issues of expression leakage, subtle expression transfer, and audio-driven precision through a carefully designed two-stage process.

In the first stage, we introduce a novel 3D Implicit Keypoints Framework that effectively disentangles motion and appearance. This stage employs a canonical loss and a landmark-guided loss to enhance the transfer of subtle expressions while simultaneously mitigating expression leakage.. These innovations significantly improve the quality and realism of generated facial animations while maintaining identity consistency. 

The second stage employs an advanced, \\audio-conditioned diffusion model. This model not only dramatically improves lip-sync accuracy but also allows for flexible control over expression and head motion parameters. By incorporating a weighted sum technique, our approach achieves unprecedented accuracy in lip synchronization, establishing a new benchmark for realistic speech-driven animations.

A key feature of Takin-ADA is its ability to generate high-resolution facial animations in real-time. Using native pytorch inference on an RTX 4090 GPU, our method achieves the generation of 512×512 resolution videos at up to 42 FPS, from audio input to final portrait output. This breakthrough in efficiency opens new possibilities for real-time digital human interaction and virtual reality applications.

Through extensive experiments and evaluations, \\we demonstrate that Takin-ADA significantly surpasses previous methods in various aspects, including video quality, facial dynamics realism, and naturalness of head movements. Our comprehensive performance enhancements not only advance the field of digital human technology but also pave the way for creating more natural and expressive AI-driven virtual characters.

In summary, Takin-ADA represents a significant step forward in single-image portrait animation, offering both technological advancements and practical applicability in real-world scenarios. By addressing the critical aspects of audio-driven avatar synthesis, our work provides a solid foundation for future research in this field and has the potential to profoundly impact various domains, including human-computer interaction, education, and entertainment.

\section{Related Work}

\subsection{3D Implicit Keypoints and Disentangled Face Representation}
The representation of facial images has been extensively studied by previous works. Traditional methods employ sparse keypoints\cite{siarohin2019first, zakharov2020fast} or 3D face models\cite{ren2021pirenderer, gao2023high, zhang2023metaportrait} to explicitly characterize facial dynamics and other properties. However, these approaches often encounter issues such as inaccurate reconstructions and limited expressive capabilities. Recent advancements have focused on learning disentangled representations within a latent space. A common strategy involves separating faces into identity and non-identity components, which are then recombined across different frames in either 2D or 3D contexts\cite{burkov2020neural, zhou2021pose, liang2022expressive, yin2022styleheat, 9578110, drobyshev2022megaportraits}. The primary challenge for these methods lies in effectively disentangling various factors while maintaining expressive representations of all static and dynamic facial attributes. Non-diffusion-based models have employed implicit keypoints as intermediate motion representations, warping the source portrait with the driving image through optical flow. Methods such as FOMM\cite{siarohin2019first} approximate local motion using first-order Taylor expansion near each keypoint and local affine transformations, whilst MRAA utilizes PCA-based motion estimation to represent articulated motion\cite{siarohin2021motion}. Face vid2vid\cite{9578110} extended the FOMM framework by introducing 3D implicit keypoints representation, achieving free-view portrait animation. Despite these advancements, Face vid2vid has limitations in the transfer of subtle expressions.

To address these challenges, several methods have been proposed to improve the warping mechanism and representation of complex motions. IWA enhanced the warping mechanism using cross-modal attention, which can be extended to multiple source images\cite{mallya2022implicit}. TPSM employed nonlinear thin-plate spline transformations to estimate optical flow more flexibly and handle large-scale motions more effectively\cite{zhao2022thin}. DaGAN leveraged dense depth maps to estimate implicit keypoints capturing critical driving movements\cite{hong2022depth} . MCNet introduced an identity representation conditioned memory compensation network to mitigate ambiguous generation caused by complex driving motions\cite{hong2023implicit}. Our work builds upon Face vid2vid\cite{9578110} by developing a series of significant enhancements to improve expression generalization and expressiveness. Our innovative use of 3D implicit keypoints forms the foundation of the Takin-ADA framework, leading to more accurate and expressive facial animations.

\subsection{Audio-Driven Talking Face Generation}
Audio-driven talking face generation has been a longstanding challenge in computer vision and graphics. Early efforts primarily focused on synthesizing lip movements from audio signals, leaving other facial attributes unchanged\cite{suwajanakorn2017synthesizing, chen2018lip, 10.1145/3394171.3413532}. Recent advancements have expanded the scope to include a broader range of facial expressions and head movements derived from audio inputs. For instance, some methods separate generation targets into categories such as lip-only 3DMM coefficients, eye blinks, and head poses, while others decompose lip and non-lip features on top of expression latents\cite{zhang2023sadtalker}. These approaches typically regress lip-related representations directly from audio features and model other attributes probabilistically\cite{yu2022talkingheadgenerationprobabilistic}. In contrast, our Takin-ADA framework generates comprehensive facial dynamics and head poses from audio along with other control signals, offering a more holistic and integrated approach to audio-driven animation.

\subsection{Diffusion Models in Facial Animation}
Diffusion models\cite{ho2020denoising} have shown remarkable performance across various generative tasks, including their application as rendering modules in facial animation\cite{du2024unicats, guo2023animatediff}. While these models often produce high-quality images, they require extensive parameters and substantial training data. To enhance generation efficiency, recent approaches have employed diffusion models for generating motion representations\cite{bigioi2024speech, he2023gaia}. Diffusion models excel at addressing the one-to-many mapping challenge crucial for speech-driven generation tasks, where the same audio clip can lead to different actions across individuals or even within the same person. The training and inference phases of diffusion models, which systematically introduce and then remove noise, allow for the incorporation of controlled variability during generation. In Takin-ADA, we leverage a state-of-the-art audio-conditioned diffusion model that integrates facial expression and head motion parameters, enabling diverse and controllable facial animations while maintaining high accuracy in lip synchronization.

\subsection{Real-Time High-Resolution Video Generation}
While recent advancements in image and video diffusion techniques have significantly improved talking face generation\cite{tian2024emo, jiang2024loopy}, their substantial computational demands have limited their practicality for interactive, real-time systems. Our work addresses this critical gap by developing a method that delivers high-quality video output while supporting real-time generation. Takin-ADA achieves the generation of 512×512 resolution videos at up to 42 FPS, from audio input to final portrait output, representing a significant advancement in the field of real-time, high-resolution facial animation.

By addressing these key areas, our Takin-ADA framework represents a comprehensive approach to audio-driven avatar synthesis, combining advanced 3D implicit keypoint representation, sophisticated audio-conditioned diffusion modeling, and efficient real-time generation capabilities.

\section{METHODOLOGY}
\begin{figure*}[ht]
    \centering
    \includegraphics[width=1\linewidth]{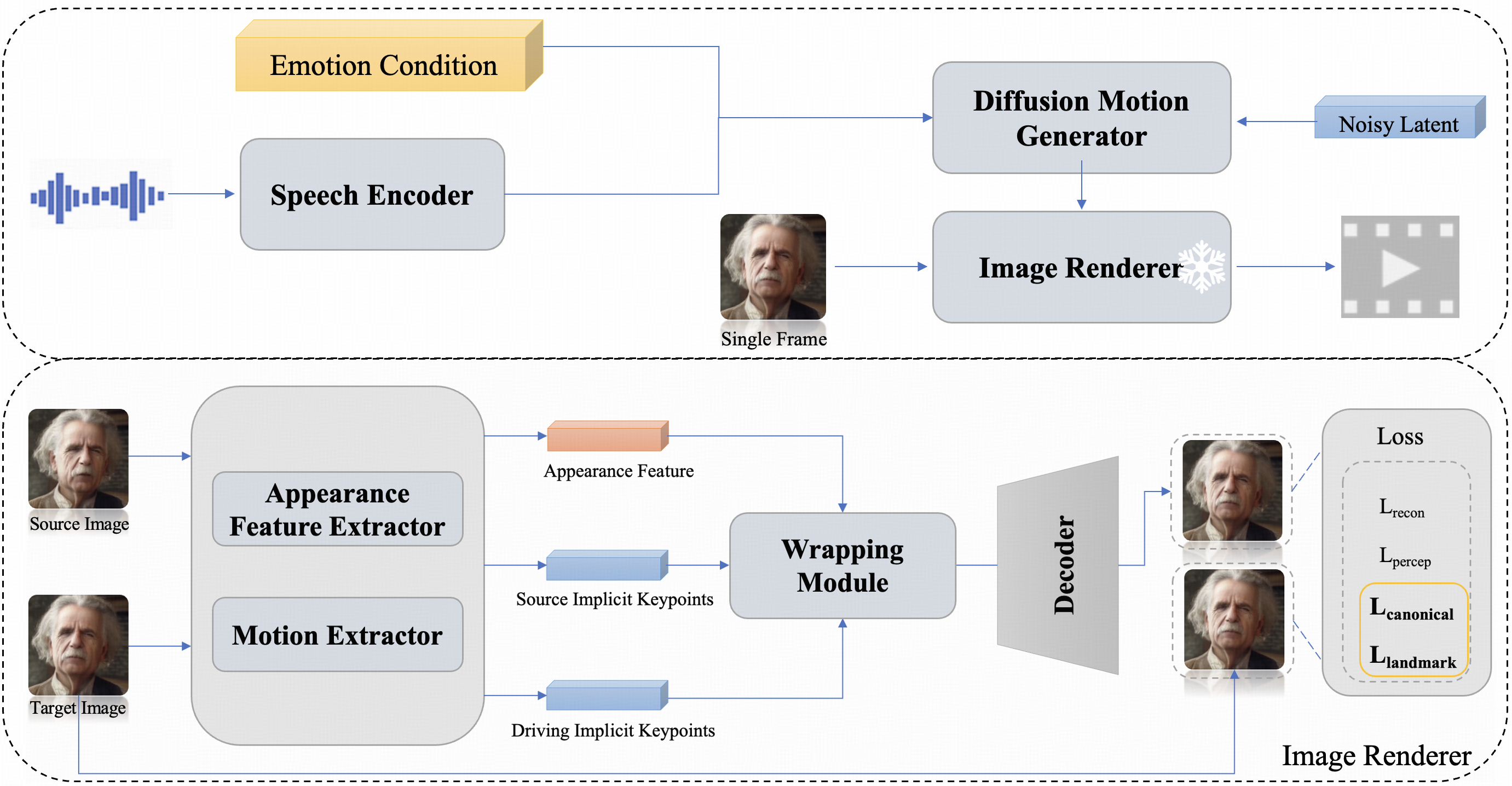}
    \caption{Illustration of our proposed Takin-ADA.
    The framework comprises two primary components: (1) a representation learning module for extracting expressive and disentangled facial latent representations, and (2) a sequence generation module that synthesizes motion sequences based on audio input.
    The first component focuses on learning robust motion representations through the utilization of canonical keypoint loss and landmark guidance. Subsequently, these learned motion representations serve as input for the second component, enabling further audio-drive facial image generation and manipulation}
    \label{fig:framework}
\end{figure*}

Figure 2 illustrates the structure of Takin-ADA, which takes a single face image of any identity and an arbitrary speech audio clip as input to generate a realistic synthesized video of the input face speaking the given audio. This section elaborates on our method in detail. We start with a brief overview of the Takin-ADA framework. Next, we describe our meticulously designed approach for constructing the latent space of the face. Finally, we introduce our comprehensive system for generating dynamic facial movements.

\subsection{Takin-ADA Framework}
Rather than directly generating video frames, we produce holistic facial dynamics and head motion in latent space, conditioned on audio and other signals. These motion latent codes are then used by a face decoder to create video frames, incorporating appearance and identity features extracted from the input image by a face encoder.  As illustrated in Figure 2, Takin-ADA encompasses two key components: 
\begin{itemize}
    \item a facial motion representation system capable of capturing universal facial dynamics. 
    \item a face latent generation using user-controlled driving signal to produce the synthesised talking face video. 
\end{itemize}

\subsection{Expressive and Disentangled Face Latent Space Construction}
In the first-stage, to build a face latent space with high degrees of expressiveness and disentanglement,  our approach utilizes a corpus of unlabeled talking face videos in a self-supervised image animation framework which employs a source image $I_s$ and a target image $I_t$ from the same video clip, where $I_s$ provides identity information, $I_t$ delivers motion details. The primary aim of our system is to reconstruct $I_t$.  
We choose face vid2vid\cite{9578110} as our base model to get facial motion latent. Compared to extant facial motion representation methodologies, including blendshapes, landmark coefficients, 2D latent and 3D Morphable Models (3DMM), the trainable latent 3D keypoints demonstrate substantial superiority in capturing nuanced emotional states and subtle facial deformations, thus providing a more sensitive and precise framework for facial animation.These 3D keypoints can be divided into two categories: one that captures facial expressions and another represents an individual's geometric signature which we called canonical volume. The 3D appearance feature volume surpassing 2D feature maps at detailing appearance. Additionally, explicit 3D feature warping proves highly effective in modeling head and facial movements in a 3D space.The source 3D keypoints $x_s$ and the driving 3D keypoints $x_d$ are transformed as follows: 
\[
 \left\{\begin{aligned}
        x_s = x_{c,s}R_s + \delta_{s} + t_s,\\
        x_d = x_{c,s}R_d + \delta_{d} + t_d,
       \end{aligned}
\ \right.
 \qquad  
\]
where $x_s$ and $x_d$ are the source and driving 3D implicit keypoints, respectively, and $x_{c,s} $represents the canonical keypoints of the source image. The source and driving poses are $R_s$ and $R_d$, the expression deformations are $\delta_{s}$ and $\delta_{d}$, and the translations are $t_s$ and $t_d$ . 

Significantly, we introduce a suite of pivotal advancements in latent 3D keypoint technology, encompassing canonical volume representation and landmark-guided optimization.

Although the canonical volume in Takin-ADA was designed to exclude facial expression details, we discovered that the generated expression is heavily influenced by the source image, indicating that information leakage affects image synthesis. To address this problem, we propose matching canonical keypoints from different images of the same person  during training, using the following loss function: 
\begin{equation}
   \mathcal{L}_{canonical} = \dfrac{1}{N}\sum_{1}^{N}(\mathcal{L}_{Huber}(x_{cs_i}, x_{cs_j}))
   \qquad 
\end{equation}
where $x_{cs_i}$ and $x_{cs_j}$are the canonical keypoints  derived from distinct images depicting the same individual. The loss serves to maintain the stability and expression-invariance of the canonical volume, which is paramount for the accurate translation of intense facial expressions. 

The original face vid2vid approach \cite{9578110} appears to have limitations in vividly animating subtle facial expressions. We posit that these shortcomings primarily stem from the inherent challenges of learning nuanced facial expressions through unsupervised methods.Drawing inspiration from \cite{guo2024liveportraitefficientportraitanimation}, we introduce 2D landmarks that capture micro-expressions, using them to guide and optimize the learning of implicit points. The landmark-guided loss $\mathcal{L}_{land}$ is formulated as follows:
\begin{equation}
   \mathcal{L}_{landmark} = \dfrac{1}{2N}\sum_{1}^{N}(\mathcal{L}_{Huber}(l_i, x_{s,i,:2}) + \mathcal{L}_{Huber}(l_i, x_{d,i,:2}))
   \qquad  
\end{equation}
where N is the number of selected landmarks, $x_{s,i,:2}$ and $x_{d,i,:2}$ denote the first two spatial dimensions of the implicit keypoints for source and driving image respectively, Huber loss is adopted following \cite{chen2017photographic}.

\textbf{Learning Objective.} The primary objective of the learning process is to reconstruct the target image by utilizing two input images: the source and the target, both belonging to the current identity index. The training process incorporates multiple loss functions to achieve this goal, including reconstruction loss $L_{recon}$, perceptual loss $L_{percep}$, Canonical Keypoints loss $L_{canonical}$, and landmark guidance loss $L_{landmark}$. These loss functions are combined to form the total loss, which is formulated as follows:
\begin{equation}
   \mathcal{L}_{total} =\lambda_1\mathcal{L}_{recon} + \lambda_2\mathcal{L}_{percep} + \lambda_3\mathcal{L}_{canonical} + \lambda_4\mathcal{L}_{landmark}
   \qquad  
\end{equation}
The values of the hyperparameters $\lambda_1$, $\lambda_2$, $\lambda_3$, and $\lambda_4$ in our experiment were chosen to be 0.1, 1, 0.2, and 0.2 respectively.

\subsection{Emotional Holistic Facial Motion Generation}
After completing the training of the motion encoder and image renderer, we freeze these models and move on to the second phase, which is driven by audio to produce motion conditioned on the audio input. Crucially, we consider holistic facial dynamics generation, where our learned latent codes represent all facial movements such as lip motion, expression, and eye gaze and blinking. Specifically, we employ a combination of diffusion and condition: the diffusion  learns a more accurate distribution of motion data, while the emotion condition primarily facilitates attribute manipulation.The trained generative model generates videos that synchronize with the speech signal or other control signals to animate a source image $I_s$.

\textbf{Diffusion formulation.} Specifically, we employ a multi-layer Conformer\cite{gulati2020conformer} for our sequence generation task. Diffusion models utilize two Markov chains: the forward chain progressively adds Gaussian noise to the target data, while the reverse chain iteratively restores the raw signal from this noise. During training, we integrate the diffusion process, where the noising phase gradually transforms clean Motion Latents M into Gaussian noise $M^T$ over a series of denoising steps. Conversely, the denoising phase systematically removes noise from the Gaussian noise\cite{ho2020denoising}, ultimately yielding clean Motion Latents. This iterative process better captures the distribution of motion, enhancing the diversity of the generated results. During the training phase, we adhere to the methodology described in \cite{ho2020denoising} for the DDPM’s training stage, applying the specified simplified loss objective, as illustrated in Equation 4, where t represents a specific time step and C represents the emotion condtion. For inference, considering the numerous iteration steps required by diffusion, we select the Denoising Diffusion Implicit Model (DDIM)\cite{song2020denoising}, an alternate non-Markovian noising process, as the solver to accelerate the sampling process.
\begin{equation}
   L_{diff} =\mathbb{E}_{t,M,\varepsilon}\lbrack \Vert \varepsilon - \hat{\varepsilon}_t(M_t, t, C) \Vert^2 \rbrack
   \qquad  
\end{equation}

\textbf{Self-Supervised Emotion Control.} To enhance facial expressions and achieve better performance, we incorporate emotional conditioning into the Conformer. Traditional multi-expression approaches typically rely on extensive multi-expression datasets, which are challenging to acquire. To address this limitation, we propose a self-supervised multi-expression generation method. Our approach is motivated by the observation that variations in facial expressions within a video sequence generally occur less frequently than other types of motion changes. Leveraging this insight, we define a window of size K around the driving image and average the K extracted expression representations to obtain a refined emotional descriptor. This refined descriptor is then combined with the extracted audio feature as input to the generator model. 

By implementing this method, we successfully achieve multi-expression capabilities without the need for explicit emotion-labeled data. Our approach not only overcomes the data acquisition hurdle but also enhances the model's ability to generate diverse and nuanced facial expressions in a self-supervised manner.During the inference phase, we can generate videos exhibiting diverse emotional states by assigning different affective vectors to the same audio input. This approach enables the production of emotionally varied outputs from a single audio source. Furthermore, we can leverage the emotional content inherent in the audio to generate videos with enhanced emotional controllability. This method allows for a more nuanced and precise manipulation of the emotional characteristics in the synthesized video output.
\section{Experiments}
\begin{table*}[ht]
    \centering
    \begin{tabular}{p{3cm}|p{2.5cm}|p{2.5cm}|p{2.5cm}|p{2.5cm}}
        \hline
        Method& Head Motion& Emotion& HD(512*512)& Real Time\\ 
        \hline
        MakeItTalk\cite{Zhou_2020} & \textcolor{red}{\ding{55}} & \textcolor{red}{\ding{55}} & \textcolor{red}{\ding{55}} & \textcolor{red}{\ding{55}} \\ \hline
        SadTalker\cite{zhang2023sadtalker} & \textcolor{blue}{\ding{51}} & \textcolor{red}{\ding{55}} & \textcolor{red}{\ding{55}} & \textcolor{red}{\ding{55}} \\ \hline
        IP\_ LAP\cite{zhang2023metaportrait}  & \textcolor{red}{\ding{55}} & \textcolor{red}{\ding{55}} & \textcolor{red}{\ding{55}} & \textcolor{red}{\ding{55}} \\ \hline
        AniTalker\cite{liu2024anitalkeranimatevividdiverse} & \textcolor{blue}{\ding{51}} & \textcolor{red}{\ding{55}} & \textcolor{red}{\ding{55}} & \textcolor{blue}{\ding{51}} \\ \hline
        EDTalk\cite{tan2024edtalkefficientdisentanglementemotional}  & \textcolor{blue}{\ding{51}} & \textcolor{blue}{\ding{51}} & \textcolor{red}{\ding{55}} & \textcolor{blue}{\ding{51}} \\ \hline
        EchoMimic\cite{echomimic}  & \textcolor{red}{\ding{55}} & \textcolor{red}{\ding{55}} & \textcolor{blue}{\ding{51}} & \textcolor{red}{\ding{55}} \\ \hline
        Takin-ADA& \textcolor{blue}{\ding{51}} & \textcolor{blue}{\ding{51}} & \textcolor{blue}{\ding{51}} & \textcolor{blue}{\ding{51}} \\ \hline
    \end{tabular}
    \caption{Summary of Different Portrait Animation Methods}
    \label{tab:comparison_methods}
\end{table*}
\subsection{Experiment Settings}
As shown in Table 1, we first give a brief summary of the key features of the existing methods.Next, we give an overview of the implementation details, dataset, benchmarks, and baselines used in the experiments. Then, we present the experimental results on video-driven methods both self-reenactment and cross-reenactment, and audio-driven methods followed by an ablation study to validate the effectiveness of the proposed calonical keypoint and landmark gudiance. 

\textbf{Implementation Details.}
The first training phase was conducted using a cluster of eight NVIDIA A800 GPUs over a 8-day period, with models initialized from scratch. Input images were preprocessed through alignment and cropping to a standardized 256×256 pixel resolution. We implemented a batch size of 104 to optimize computational efficiency, while the output resolution was set at 512×512 pixels. We follow \textit{Face Vid2Vid} \cite{9578110} to use implicit keypoints equivariance loss $\mathcal{L}_{E}$, keypoint prior loss $\mathcal{L}_{L}$, head pose loss $\mathcal{L}_{H}$, and deformation prior loss $\mathcal{L}_{\Delta}$. To further improve the expression disentanglement, we apply Canonical Keypoints losses and Landmark Guidance losses, denoted as $\mathcal{L}_{\text{canonical}}$ and $\mathcal{L}_{\text{landmark}}$. To further improve the texture quality, we also apply perceptual and GAN losses on the global region of the input image fine-tuned from \textit{LivePortrait} model. In the second phase, the speech encoder and the Motion Generator utilize a four-layer and an eight-layer conformer architecture, respectively, inspired by \cite{du2023dae}. This architecture integrates the conformer structure and relative positional encoding \cite{dai2019transformer, gulati2020conformer}. A pre-trained HuBERT-large model \cite{hsu2021hubert} serves as the audio feature encoder, incorporating a downsampling layer to adjust the audio sampling rate from 50 Hz to 25 Hz to synchronize with the video frame rate. The training of the audio generation process spans 125 frames (5 seconds). To enhance the robustness of the audio encoder, we employ a novel approach that retrieves the audio latent code through a weighted summation of all layers within the self-supervised models. 


\textbf{Dataset.} Our study employs three distinct datasets: VoxCeleb\cite{nagrani2017voxceleb}, HDTF\cite{zhang2021flow}, and MEAD\cite{wang2020mead}. To ensure consistency in data processing, we retrieved the original video files from these sources and implemented a standardized processing methodology across all datasets. Furthermore, we augmented our research with a substantial collection of 4K-resolution portrait videos, comprising approximately 200 hours of talking head footage.
In preprocessing this additional data, we segmented extended video sequences into clips not exceeding 30 seconds in duration. To maintain data integrity and focus, we utilized face tracking and recognition technologies to ensure that each clip contains footage of only a single individual. This approach enhances the dataset's suitability for our research objectives and facilitates more accurate analysis.

\textbf{Benchmarks.} To quantitatively measure the visual quality, we figure up the Peak Signal-to-Noise Ratio (PSNR), Structure SIMilarity (SSIM) and Learned Perceptual Image Patch Similarity (LPIPS) for the generated videos\cite{wang2004image, zhang2018unreasonable}. Following Wav2Lip\cite{10.1145/3394171.3413532}, Lip-sync Distance (LSE-D) is applied to measure the audiovisual synchronization. For assessing reenactment quality, we employ various metrics including the Frechet Inception Distance (FID) to measure the distributional discrepancy between synthetic and real images\cite{heusel2017gans}. Cosine similarity (CSIM) from a face recognition network quantifies the identity preservation in generated images\cite{cao2018vggface2} and Structural Similarity Index (SSIM)\cite{SSIM} . Regarding subjective metrics, we employ the Mean Opinion Score (MOS) as our metric, with 35 participants rating our method based on Lip-sync(LS), Naturalness(N), Resolotion(R), and Expression Transfer(ET) .

\subsection{Summary of the portrait animation methods}
Table 1 summarizes the key features of existing methods in terms of high-quality output (HD), real-time performance, and fine-grained control over different aspects, including head motion and emotion. While other approaches excel in some areas, our method uniquely possesses all these desirable characteristics. This comprehensive capability is made possible by our sophisticated universal motion representation, which enables us to balance quality, efficiency, and control effectively. Our approach thus represents a significant advancement in speech-driven facial animation technology, offering a solution that doesn't compromise on any front.

\subsection{Video-driven methods}
\begin{table*}[ht]
    \centering
    \begin{tabular}{l|c|c|c|c|c|c|c|c}
        \hline
        \multirow{2}{*}{Method} & \multicolumn{4}{c|}{Self-Reenactment} & \multicolumn{3}{c}{Cross-Reenactment} \\ \cline{2-8} 
         & FID$\downarrow$ & CSIM$\uparrow$ & LPIPS$\downarrow$ & MOS-ET$\uparrow$ & CSIM$\uparrow$ & LPIPS$\downarrow$ & MOS-ET$\uparrow$ \\ \hline
        FOMM\cite{siarohin2019first} &  32.935&  0.825&  0.021&  2.769&  0.474&  0.418&  1.934\\ \hline
        StyleHEAT\cite{yin2022styleheat} &  33.136&  0.522&  0.095&  2.675&  0.544&  0.413&  1.768\\ \hline
        LIA\cite{wang2022latent} &  28.008&  0.834&  0.021&  3.187&  0.449&  0.416&  2.937\\ \hline
        FADM\cite{zeng2023face} &  28.981&  0.832&  0.024&  2.763&  0.406&  0.411&  2.268\\ \hline
        Face Vid2Vid\cite{9578110} &  28.444&  0.831&  0.023&  3.451&  0.444&  0.412&  2.664\\ \hline
        Takin-ADA &  \textbf{27.429}&  \textbf{0.948}&  \textbf{0.019}&  \textbf{3.983}&  \textbf{0.561}&  \textbf{0.399}&  \textbf{3.575}\\ \hline
    \end{tabular}
    \caption{Quantitative comparisons for self-reenactment and cross-reenactment methods.}
    \label{tab:self_cross_reenactment_metrics}
\end{table*}
\begin{figure*}[ht]
    \centering
    \includegraphics[width=1\linewidth]{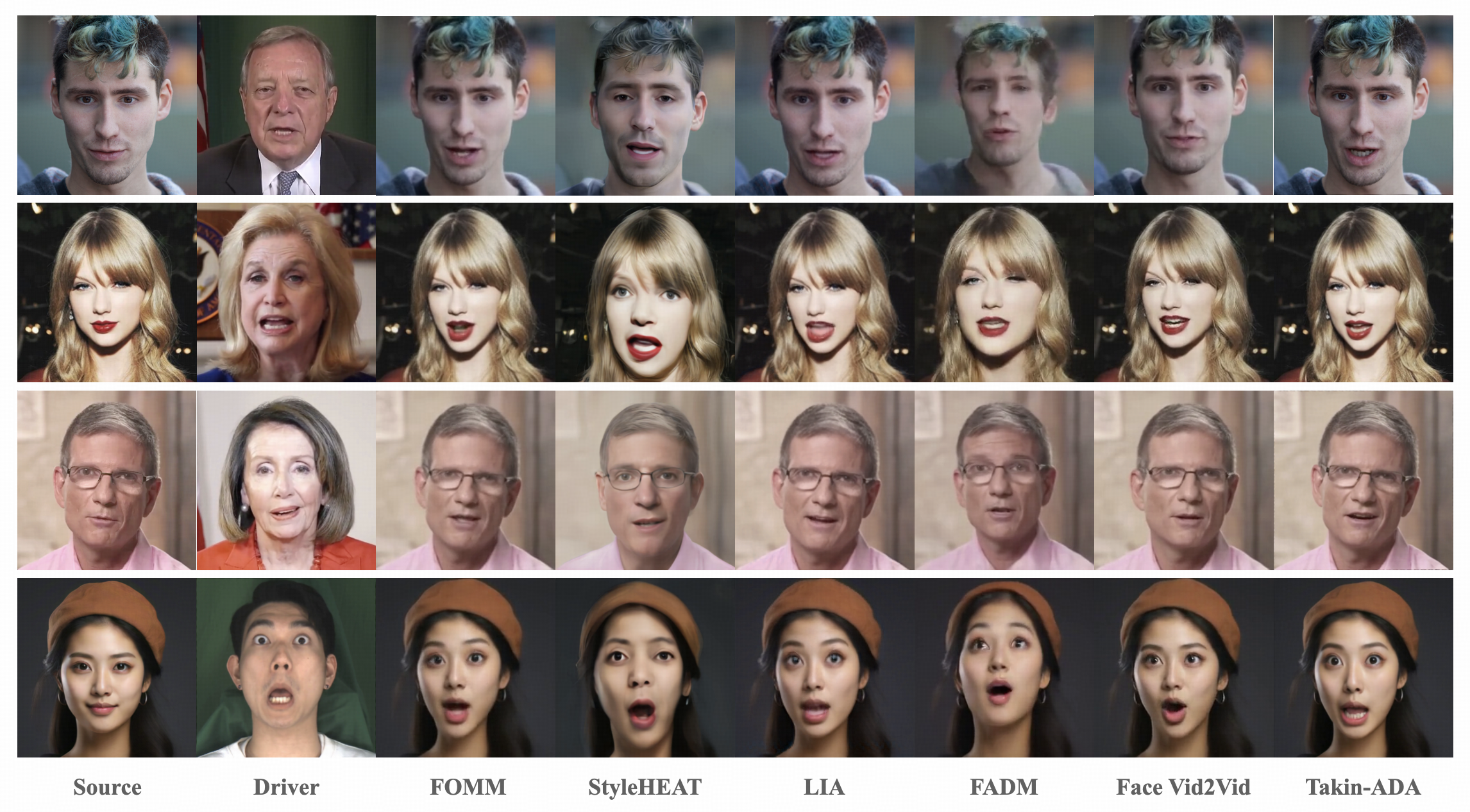}
\caption{Qualitative comparisons of Cross-reenactment. This task involves transferring actions from a source portrait to a target portrait to evaluate each algorithm's ability to separate motion and appearance. The results highlight our method's superior ability in both motion transfer and appearance retention, while also excelling in the transfer of subtle micro-expressions and extreme facial expressions.}
    \label{fig:qualitative comparisons}
\end{figure*}
\textbf{Quantitative Results.} We benchmarked our approach against several leading face reenactment methods, all employing variations of self-supervised learning. The results are presented in Table 1. Due to the inherent challenges and the absence of frame-by-frame ground truth in Cross-Reenactment (using another person’s video for driving), the overall results tend to be lower compared to Self-Reenactment (using the current person’s video). In Self-Reenactment, our algorithm achieved superior results for image structural metrics such as FID, CSIM, and LPIPS, validating the effectiveness of our motion representation in reconstructing images. Specifically, Takin-ADA achieved a FID score of 27.429, which is notably lower than FOMM and Vid2Vid, indicating a smaller distributional discrepancy between generated and real images. Additionally, the CSIM score of 0.937 surpasses other methods, demonstrating better identity preservation. The lowest LPIPS value of 0.019 further confirms the superior visual quality of our generated results. In the cross-reenactment task, our method also shows significant advantages, especially in terms of CSIM and LPIPS metrics. Our system effectively separates the driving actions and identity features, retaining the target head movements and expressions while preserving the source identity. The high MOS-ET score also reflects the high subjective satisfaction with our method. Takin-ADA achieved the best performance among all methods, with a CSIM score of 0.561 and a MOS-ET score of 0.375.   These results highlight our algorithm's outstanding ability to disentangle identity and motion when driving with different individuals, providing more natural, expressive, and high-fidelity facial animations.

\begin{table*}[ht]
    \centering
    \begin{tabular}{l|c|c|c|c|c|c|c}
        \hline
        \multirow{2}{*}{Method} & \multicolumn{3}{c|}{Subjective Evaluation} & \multicolumn{4}{c}{Objective Evaluation} \\ \cline{2-8} 
         & MOS-R$\uparrow$& MOS-N$\uparrow$& MOS-LS$\uparrow$& PSNR$\uparrow$ & SSIM$\uparrow$ & FID$\downarrow$ & LSE-D$\downarrow$  \\ \hline
        MakeItTalk\cite{Zhou_2020} &  2.135&  2.822&  2.441&  26.693&  0.762&  31.113& 10.888\\ \hline
        SadTalker\cite{zhang2023sadtalker} &  3.783&  2.148&  3.573&  26.105&  0.753&  32.539& 7.748\\ \hline
        AniPortrait\cite{AniPortrait} &  3.529&  2.329&  3.474&  25.172&  0.731&  33.434& 7.968\\ \hline
        AniTalker\cite{liu2024anitalkeranimatevividdiverse} &  3.956&  2.812&  3.821&  25.387&  0.749&  29.839& 10.171\\ \hline
        EDTalk\cite{tan2024edtalkefficientdisentanglementemotional}  &  2.943&  3.152&  3.752&  26.978&  \textbf{0.781}&  28.043& \textbf{7.686}\\ \hline
        Takin-ADA&  \textbf{4.187}&  \textbf{3.839}& \textbf{3.887}&  \textbf{27.876}&  0.779&  \textbf{27.803}& 7.764\\ \hline
    \end{tabular}
    \caption{Quantitative comparisons with previous speech-driven methods.}
    \label{tab:evaluation_metrics}
\end{table*}
\textbf{Qualitative Results.} Figure 3 presents a qualitative comparison of cross-reenactment methods. This task involves transferring actions from a source portrait to a target portrait to evaluate each algorithm's ability to separate motion and appearance. From the third row, it is clear that our method, Takin-ADA, excels in transferring subtle micro-expressions, effectively capturing and replicating delicate facial movements. From the fourth row, Takin-ADA also shows superior performance in handling extreme facial expressions, maintaining the integrity and authenticity of the facial features even under challenging conditions. These results highlight the robustness and effectiveness of Takin-ADA in both subtle and extreme expression transfer.

\subsection{Audio-driven methods}

To assess the efficacy of our proposed method, we conducted a comprehensive evaluation comparing it against state-of-the-art speech-driven approaches. The comparison encompassed both subjective and objective evaluations, including benchmarking against notable works such as MakeItTalk\cite{Zhou_2020}, SadTalker\cite{zhang2023sadtalker}, AniPortrait\cite{AniPortrait}, AniTalker\cite{liu2024anitalkeranimatevividdiverse}, and EDTalk\cite{tan2024edtalkefficientdisentanglementemotional}. Table 3 presents a detailed quantitative analysis of the comparative results.

Subjective evaluations consistently demonstrate that our method outperforms existing techniques in lip-sync accuracy (MOS-LS), naturalness (MOS-N),and Resolution (MOS-R), with particular emphasis on enhanced naturalness of movements. These improvements can be attributed to our sophisticated universal motion representation. Notably, our model demonstrates a superior ability to produce convincingly synchronized lip movements that accurately match the given phonetic sounds. Nevertheless, our SSIM\cite{SSIM} and LSE-D metric exhibits a slight decline compared to EDTalk, which we attribute to two primary factors: 1) EDTalk \cite{tan2024edtalkefficientdisentanglementemotional} is exclusively trained on lip movements, whereas our model predicts the full range of facial expressions.  2) the LSE-D metric emphasizes short-term alignment, 3) the metric is not utilized as a supervisory signal in our training process, thereby failing to sufficiently capture the long-term information essential for the comprehensibility of generated videos. This observation is further supported by the qualitative results presented in Figure 4, which underscore our model's capability to produce convincingly synchronized lip movements corresponding to the provided phonetic sounds.

\textbf{Consistency with the provided phonetic sounds.} Figure 4 demonstrates our model's proficiency in generating highly synchronized lip movements that correspond accurately to the given phonetic sounds. This visual representation underscores the model's capability to create realistic and precisely timed facial animations that align seamlessly with spoken language.
\begin{figure}[!h]
    \centering
    \includegraphics[width=1.0\linewidth]{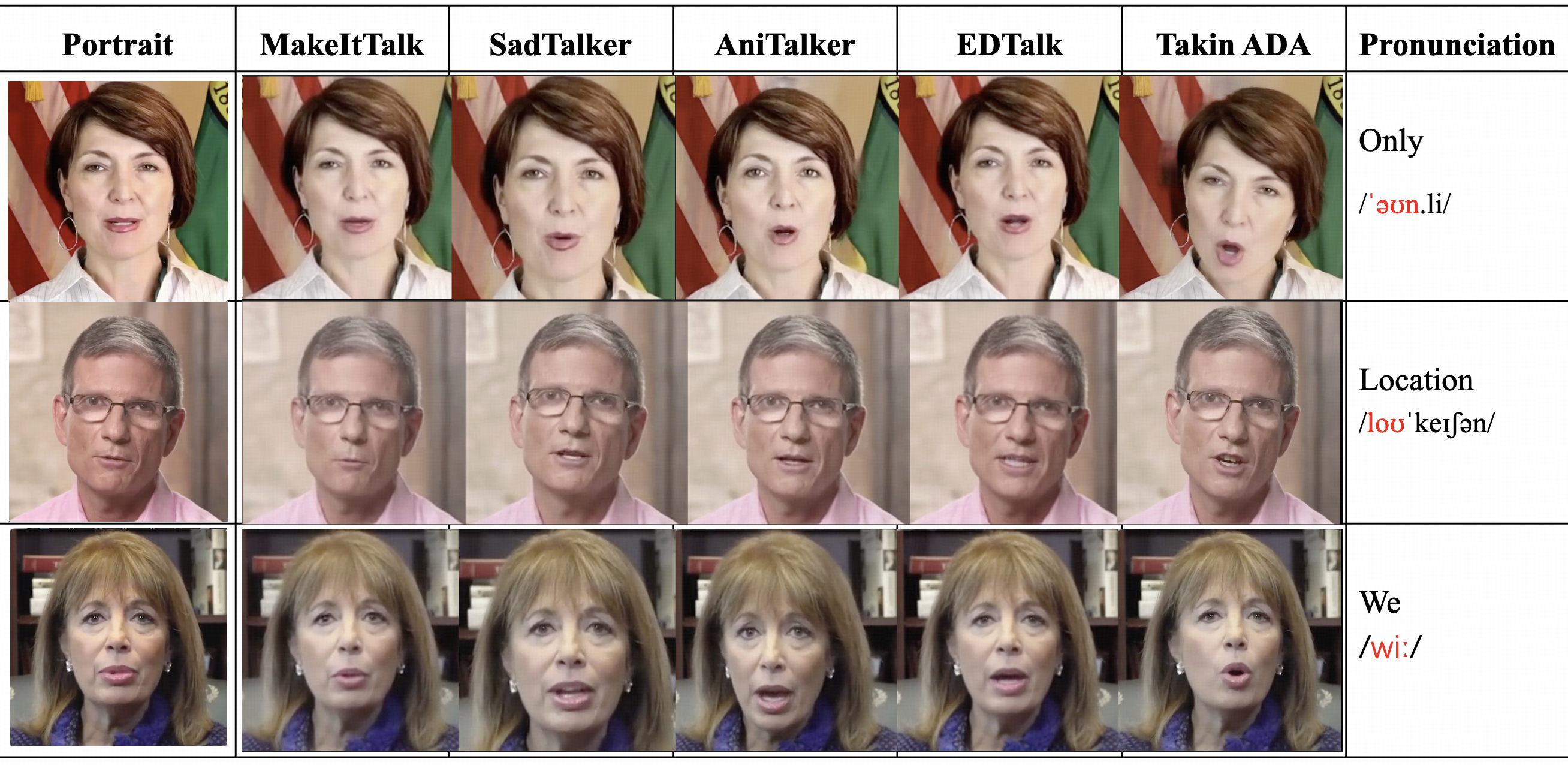}
    \caption{Visual comparison of the speech-driven method. Phonetic sounds are highlighted in red.}
    \label{fig:visual comparison}
\end{figure}

\textbf{Emotion Control.} Figure 5 presents a diverse array of our generated results, encompassing various emotional states. The emotion condition is extract from emotional image using stage 1, These examples vividly demonstrate our generation model's proficiency in interpreting emotional signals and producing talking face animations that closely correspond to the specified emotional parameters.

\begin{figure}[!h]\textit{}
    \centering
    \includegraphics[width=1\linewidth]{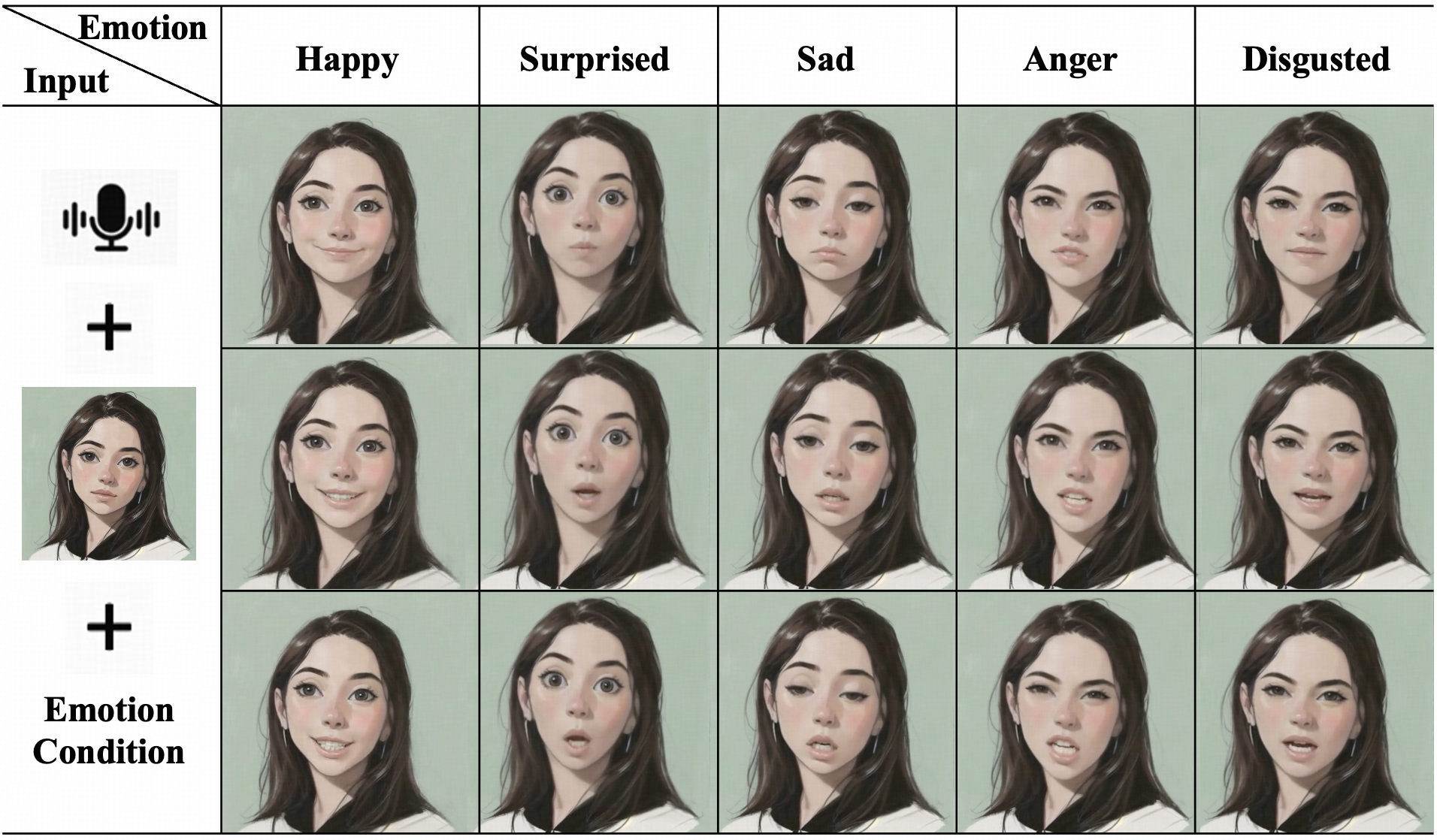}
    \caption{Generated results under different emotion offset (happy, surprised, sad, angry and disgusted, respectively). }
    \label{fig:emotion results}
\end{figure}

The results unequivocally showcase the model's capacity to accurately capture and convey a wide spectrum of emotions through the generated facial expressions and movements. This underscores the system's effectiveness in translating emotional inputs into visually convincing and emotionally resonant animations.

\subsection{Ablation Study}
To rigorously evaluate and substantiate the efficacy of our proposed canonical and landmark guidance methodology, we conducted a comprehensive ablation study encompassing multiple experimental configurations. First, to evaluate the performance of our model without the canonical loss ($\mathcal{L}_{canonical}$), we observed the resulting metrics and compared them against a fine-tuned vid2vid baseline. This comparison, detailed in Table 4, demonstrates significant improvements across all metrics when either component is added. The exclusion of $\mathcal{L}_{canonical}$ resulted in moderate improvements, with a MOS-ET of 3.542, CSIM of 0.947. However, the lack of $\mathcal{L}_{canonical}$ primarily impacts the ability to translate extreme expressions, as indicated by MOS-ET. In contrast, incorporating $\mathcal{L}_{canonical}$ can significantly reduce the effect of the source expression, ensuring the results are stable and expression-independent, which enhances the translation of intense expressions and reduces expression leakage. The exclusion of $\mathcal{L}_{landmark}$ yielded better results. $\mathcal{L}_{landmark}$ is crucial for accurately transferring subtle facial expressions, leading to a better performance in capturing micro-expressions. By incorporating both $\mathcal{L}_{canonical}$ and $\mathcal{L}_{landmark}$, our complete method achieved the best results. These results highlight the powerful synergy of these disentanglement losses, leading to enhancements in image quality, structural similarity, and expression transfer.


\begin{table}[ht]
    \footnotesize
    \centering
    \begin{tabular}{c|c|c|c|c}
        \hline
        Method& FID$\downarrow$& CSIM$\uparrow$& MOS-ET$\uparrow$& PSNR$\uparrow$\\ 
        \hline
        Face Vid2Vid fine-tuned& 28.444& 0.945& 3.451& 19.235\\ 
        \hline
        Ours w/o $\mathcal{L}_{\text{canonical}}$& 28.721& 0.947& 3.542& 22.254\\ 
        Ours w/o $\mathcal{L}_{\text{landmark}}$& 27.828& 0.948& 3.662& 23.619\\ 
        \hline
        Ours & \textbf{27.429}& \textbf{0.948}& \textbf{3.983}& \textbf{24.663}\\ 
        \hline
    \end{tabular}
    \caption{Quantitative comparisons of disentanglement methods in Self-Reenactment setting}
    \label{tab:method_comparison}
\end{table}



\section{CONCLUSIONS}
\label{sec:CONCLUSIONs}

We introduced Takin-ADA, an innovative two-stage\\ framework for real-time audio-driven animation of single-image portraits with controllable expressions using 3D implicit keypoints. Our approach addresses critical limitations in existing methods, including expression leakage, subtle expression transfer, and lip-sync accuracy. By employing a canonical loss and a landmark-guided loss in the first stage, Takin-ADA enhances the transfer of subtle expressions and reduces expression leakage. In the second stage, an audio-conditioned diffusion model significantly improves lip-sync accuracy. Additionally, Takin-ADA achieves high-resolution (512×512) facial animations at up to 42 FPS on an RTX 4090 GPU. Extensive evaluations demonstrate that Takin-ADA consistently outperforms existing solutions in video quality, facial dynamics realism, and naturalness of head movements.

While Takin-ADA shows significant advancements, it still has some limitations, such as minor inconsistencies in complex backgrounds and edge blurring during extreme facial shifts. Our future work will focus on improving the temporal coherence and rendering quality of the framework. Takin-ADA sets a new benchmark in single-image portrait animation, opening new possibilities for applications like virtual hosts, online education, and digital human interactions, and providing a robust foundation for future research in this evolving field.



{\small
\bibliographystyle{cvm}
\bibliography{cvmbib}
}

\end{document}